\documentclass[10pt,twocolumn,letterpaper]{article}
\usepackage[T1]{fontenc}
\usepackage[pagenumbers]{cvpr}

\usepackage{cuted} 

\usepackage{currfile} 

\usepackage{caption} 

\AtBeginDocument{\hfuzz=0.5pt\vfuzz=0.5pt}
\usepackage{mathtools}
\usepackage{tabularx,array,multirow}
\usepackage{tikz}
\usetikzlibrary{arrows.meta,positioning,calc}
\usepackage{float}
\usepackage{placeins}
\newcolumntype{Y}{>{\raggedright\arraybackslash}X}

\floatstyle{ruled}
\newfloat{algorithm}{tbp}{loa}
\floatname{algorithm}{Algorithm}
\newcounter{suppalgline}
\newlength{\suppalgindent}
\newenvironment{suppalgorithmic}{%
  \setcounter{suppalgline}{0}\setlength{\suppalgindent}{0pt}%
  \begin{list}{}{\setlength{\leftmargin}{2.1em}%
    \setlength{\labelwidth}{1.6em}\setlength{\labelsep}{0.5em}%
    \setlength{\itemsep}{1pt}\setlength{\parsep}{0pt}%
    \setlength{\topsep}{3pt}\setlength{\partopsep}{0pt}}%
}{\end{list}}
\newcommand{\AlgState}{\refstepcounter{suppalgline}%
  \item[\scriptsize\arabic{suppalgline}:]\hspace*{\suppalgindent}}
\newcommand{\AlgRequire}{\item[]\textbf{Require:} }
\newcommand{\AlgEnsure}{\item[]\textbf{Ensure:} }
\newcommand{\AlgIf}[1]{\AlgState\textbf{if} #1 \textbf{then}\addtolength{\suppalgindent}{1.5em}}
\newcommand{\AlgElse}{\addtolength{\suppalgindent}{-1.5em}\AlgState\textbf{else}\addtolength{\suppalgindent}{1.5em}}
\newcommand{\AlgEndIf}{\addtolength{\suppalgindent}{-1.5em}\AlgState\textbf{end if}}
\newcommand{\AlgWhile}[1]{\AlgState\textbf{while} #1 \textbf{do}\addtolength{\suppalgindent}{1.5em}}
\newcommand{\AlgEndWhile}{\addtolength{\suppalgindent}{-1.5em}\AlgState\textbf{end while}}
\newcommand{\AlgReturn}{\textbf{return} }

\newcommand{\E}{\mathbb{E}}
\newcommand{\KL}{D_{\mathrm{KL}}}
\newcommand{\TC}{\operatorname{TC}}
\newcommand{\bpsp}{\mathrm{bpsp}}
\newcommand{\bpp}{\mathrm{bpp}}
\newcommand{\BR}{\mathrm{BR}}

\definecolor{cvprblue}{rgb}{0.21,0.49,0.74}
\usepackage[breaklinks,colorlinks,allcolors=cvprblue]{hyperref}
\hypersetup{pdftitle={Entropy Is Invariant, Codelength Is Not: Representation and Conditional Factorization in Neural Lossless Image Compression}}
\def\confName{CVPR}
\def\confYear{2027}

\title{Entropy Is Invariant, Codelength Is Not: Representation and Conditional Factorization in Neural Lossless Image Compression}
\author{
  ZiGuang Li$^{1,2,*}$\quad
  Lang Zheng$^{1,2,*}$\quad
  Yan Xu$^{1,2,*}$\quad
  Ming Li$^{3,\dagger}$\\[4pt]
  $^{1}$Fudan University\\
  $^{2}$Shanghai Institute for Mathematics and Interdisciplinary Sciences\\
  $^{3}$University of Waterloo\\[4pt]
  {\small $^{*}$Equal contribution.\quad $^{\dagger}$Corresponding author.}
}
\hypersetup{pdfauthor={ZiGuang Li, Lang Zheng, Yan Xu, Ming Li}}

\begin{document}
\maketitle
\begin{abstract}
The coding efficiency of neural lossless image compression is determined not only by the architecture of the probability model, but also by the prediction problem induced by the representation and conditional factorization. We investigate how invertible representations can improve coding efficiency while preserving image information. Our analysis decomposes expected ideal residual codelength into conditional entropy and two excess terms arising from independent within-group prediction and imperfect model fitting. Guided by this perspective, we develop a base-guided progressive residual codec that combines integer-reversible color decorrelation with modular residuals relative to a transmitted low-resolution base. A deterministic block-raster plan organizes residual recovery into successive groups, exposing exact decoded values as context alongside multi-scale base features. Training directly targets the conditional distributions required by this recovery process, and the resulting codec reconstructs every RGB value exactly. On a controlled 256-image development set, color-residual coordinates reduce complete-file rate by 8.71\% relative to direct RGB prediction under the
same base, grouping, and training protocol. Evaluation on 226 original-resolution images shows lower complete-file rates than JPEG XL at effort 11
on DIV2K and both CLIC2020 subsets. These results connect information-preserving representation design and conditional organization to measurable reductions in actual coding cost.

\end{abstract}

\section{Introduction}
\label{sec:intro}

Recent advances in neural lossless image compression have improved coding efficiency through more expressive probability models and more effective exploitation of image dependencies. Existing approaches have explored complementary mechanisms, including hierarchical probability modeling~\cite{mentzer2019l3c}, reference-guided residual coding~\cite{bai2024dlpr,mentzer2020lossy,li2025masked}, reversible transformations~\cite{hoogeboom2019idf}, structured conditional prediction~\cite{rhee2022lcfdnet,fu2024hybrid}, and image-adaptive probability modeling~\cite{zhang2025fnlic}. Traditional codecs such as FLIF and JPEG XL have likewise exploited reversible transformations, prediction, and adaptive context modeling~\cite{sneyers2016flif,alakuijala2019jpegxl}. Despite these advances, how representation choices and decoder-available conditional information jointly shape the prediction tasks faced by practical neural probability models---and ultimately their achievable coding efficiency---remains insufficiently understood. In particular, the benefits of individual design choices need not be independent or additive, as their effectiveness depends on the conditional structure exposed during decoding.

An invertible transformation preserves the information contained in an image, yet can substantially reorganize its statistical dependencies and alter the difficulty of the resulting prediction tasks. For a bijective transformation $T$, the joint entropy remains invariant, $H(X)=H(T(X))$~\cite{cover2006elements}. However, this invariance does not imply that a practical neural probability model achieves the same coding length across different representations. The model must estimate the conditional distributions induced by a particular representation and factorization, and its achievable cross-entropy depends on how effectively it captures the statistical regularities exposed by them~\cite{cover2006elements,kullback1951information}. Consequently, two information-equivalent representations can lead to substantially different model-achievable codelengths, even though neither contains more information than the other. This distinction separates information-theoretic entropy from practical neural coding efficiency: the latter is determined not only by the information to be encoded, but also by how favorably that information is organized for probability estimation. To make this distinction operational, we consider the complete-file coding cost
\[
L_{\mathrm{file}}
=
L_{\mathrm{base}}
+
L_{\mathrm{res}}(T,\pi,\theta)
+
L_{\mathrm{overhead}},
\]
where $L_{\mathrm{base}}$, $L_{\mathrm{res}}$, and $L_{\mathrm{overhead}}$ denote the base payload, residual bitstream, and coding overhead, respectively; $T$, $\pi$, and $\theta$ represent the residual representation, decoding schedule, and neural probability model. This perspective motivates designing representations not merely to preserve information or reduce residual magnitudes, but to make the resulting symbols more predictable under practical neural probability estimation.

Representation alone, however, does not determine the conditional information available during actual decoding. In a group-wise parallel codec, the decoding order determines which symbols have already been recovered and can therefore serve as exact conditioning information. At each stage, symbols within the current group must be predicted without access to the unknown ground-truth values of other unrecovered symbols, while previously decoded groups can provide additional context~\cite{reed2017parallel,he2021checkerboard,he2022elic}. Consequently, statistical dependencies induced by a representation cannot necessarily be exploited unless the corresponding information is accessible through the decoding process. Conversely, an appropriate decoding organization can expose useful dependencies that would otherwise remain unavailable to the predictor. The decoding order is therefore not merely an implementation choice for balancing parallelism and sequential prediction; it defines the decoder-realizable conditional structure under which the probability model operates. Together, these observations motivate the joint design of reversible representations and decoding-induced conditional structures to construct prediction tasks that are more favorable for practical neural probability models.

\begin{figure*}[t]
    \centering
    \includegraphics[
        width=0.9\textwidth,
        height=0.3\textheight,
        keepaspectratio
    ]{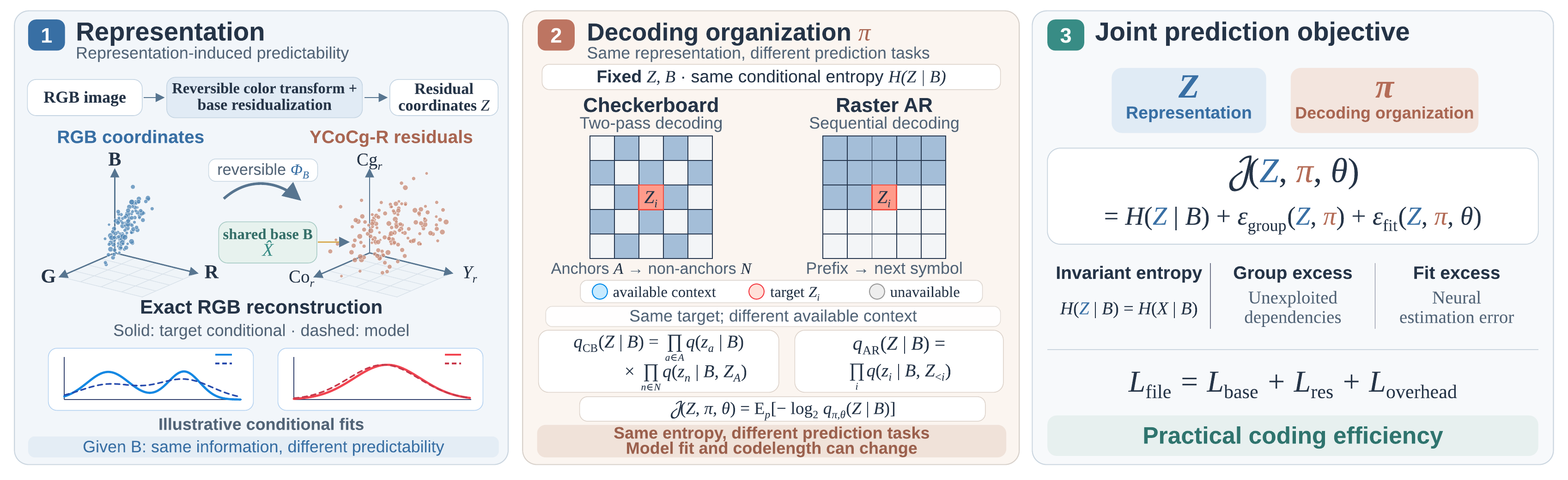}

    \caption{
    Conceptual overview of how reversible representations and
    decoder-available context shape practical coding efficiency.
    (Left) Reversible color transformation and base-conditioned
    residualization preserve exact recovery while changing
    model-facing predictability.
    (Right) For fixed $Z$ and $B$, different decoding organizations,
    such as checkerboard and raster autoregressive decoding,
    expose different contexts and induce different prediction tasks.
    Together, these choices affect the grouped-prediction and
    model-fitting excesses above the invariant conditional entropy,
    thereby shaping the complete-file codelength.
    }
    \label{fig:conceptual-overview}
\end{figure*}
Based on this perspective, we develop a reversible color-residual representation that combines integer-reversible YCoCg-R transformation~\cite{malvar2003ycocgr,malvar2008ycocg} with base-guided residual coding~\cite{mentzer2020lossy,bai2024dlpr}. We first encode a compact base representation, which is deterministically reconstructed at both the encoder and decoder, and then encode the exact integer residuals in the transformed color space. This design reorganizes cross-channel dependencies and shifts prediction from absolute image values to information not captured by the base reconstruction. The shared reconstruction also provides spatial conditioning without additional side information, while its transmission cost is accounted for in the complete-file coding objective.

To exploit dependencies among residual symbols, we introduce a block-raster causal grouping strategy that progressively exposes previously decoded residuals as conditioning information while preserving parallel prediction within each group. We employ a discrete residual diffusion model for probability estimation, with its training process aligned with the decoding structure. This enables the model to exploit decoder-available context through a structured conditional prediction process, with entropy coding ensuring exact recovery~\cite{rissanen1979arithmetic,witten1987arithmetic}.

Our contributions are summarized as follows:
\begin{itemize}
    \item \textbf{Reversible color-residual representation.}
    We develop a reversible representation that combines integer-reversible YCoCg-R transformation with base-guided residual modeling. Through controlled ablations, we quantify the individual and combined effects of cross-channel decorrelation and residual recentering, revealing non-additive coding gains and highlighting the importance of evaluating their interactions in terms of complete-file coding cost.

    \item \textbf{Decoding-induced conditional structure.}
    We introduce a block-raster causal grouping strategy that organizes residual symbols into parallel prediction groups while progressively exposing previously recovered symbols as exact conditioning information. Controlled decoding-order comparisons demonstrate its coding advantage across models trained with different decoding strategies, showing that the benefit extends beyond training-inference schedule matching.

    \item \textbf{End-to-end lossless codec and evaluation.}
    We integrate the proposed representation and decoding
strategy into a discrete residual diffusion codec,
enabling decoder-aligned probability estimation and
exact image reconstruction. We evaluate complete-file
coding costs, including base payloads and coding
overhead, on four original-resolution test sets,
demonstrating competitive rates against established
lossless codecs.
\end{itemize}

\section{Related Work}
\label{sec:related}

\subsection{Learned Lossless Compression and Reversible Representations}

Learned lossless image compression has primarily improved coding efficiency through more expressive probability models and structured probabilistic factorizations. L3C~\cite{mentzer2019l3c} employs hierarchical probabilistic modeling with learned auxiliary representations, while SReC~\cite{cao2020srec} progressively models images across spatial scales. Integer discrete flows~\cite{hoogeboom2019idf} learn bijective mappings for probability estimation, and other approaches investigate latent-variable coding~\cite{townsend2019bitsback} and image-adaptive probability estimation~\cite{zhang2025fnlic}. Li et al.~\cite{li2025masked} combine a lossy reconstruction with residual probability estimation conditioned on previously decoded symbols, using prediction-dependent masks to select symbols for coding.

Another line of work reorganizes image information through reversible representations. Integer-reversible color transformations, such as YCoCg-R~\cite{malvar2008ycocg}, use lifting operations to transform RGB values into luminance and chrominance components while preserving exact invertibility. Other approaches reorganize image symbols through bit-plane decomposition~\cite{zhang2024bps}, frequency-aware subimage coding~\cite{rhee2022lcfdnet}, and alternative color spaces with multi-scale context modeling~\cite{fu2024hybrid}.

These approaches highlight two complementary aspects of practical neural lossless compression: the expressive capability of the probability model and the statistical properties of the representation it operates on. Although a bijective transformation preserves joint entropy, $H(X)=H(T(X))$~\cite{cover2006elements}, it does not guarantee equivalent prediction difficulty for a practical neural probability model. Different representations can induce different conditional distributions and prediction tasks, resulting in different achievable cross-entropies even under the same model architecture and capacity~\cite{cover2006elements,kullback1951information}. Thus, effective entropy modeling depends not only on model expressiveness but also on the predictability induced by the input representation. While existing codecs have incorporated various reversible transformations into their coding pipelines, how representation-induced predictability shapes practical neural coding efficiency remains insufficiently characterized.

\subsection{Grouped Context Models}

Autoregressive models such as PixelRNN~\cite{oord2016pixelrnn} and PixelCNN++~\cite{salimans2017pixelcnnpp} factorize image distributions by conditioning predictions on previously decoded values. To enable parallel prediction, multiscale autoregressive density estimation~\cite{reed2017parallel} organizes symbols into conditionally independent groups. Related strategies appear in learned codecs: L3C~\cite{mentzer2019l3c} predicts spatial positions in parallel within each scale, SReC~\cite{cao2020srec} progressively reconstructs spatial resolutions, and PILC~\cite{kang2022pilc} combines autoregressive modeling with parallel entropy coding. HPAC~\cite{li2025hpac} further explores hierarchical autoregressive factorization with computation reuse. In lossy compression, checkerboard context modeling~\cite{he2021checkerboard} predicts non-anchor latents using decoded anchors.

These approaches establish different ways of organizing conditional dependencies while balancing parallelism and sequential context utilization~\cite{minnen2020channel,he2022elic}. In practical group-wise decoding, however, the available conditional information is determined by both the symbol representation and the decoding schedule. A reversible transformation can reorganize dependencies among symbols, while the grouping and decoding order determine which of these dependencies become accessible as exact context at each stage. Thus, the usefulness of a conditional factorization cannot be considered independently of the representation being encoded. How representation-induced dependencies interact with decoder-available conditional information, and how this interaction shapes the prediction tasks of practical probability models, remains an important question.

\section{Method}
\label{sec:method}

\begin{figure*}[t]
    \centering
    \includegraphics[
        width=\textwidth,
        height=0.35\textheight,
        keepaspectratio
    ]{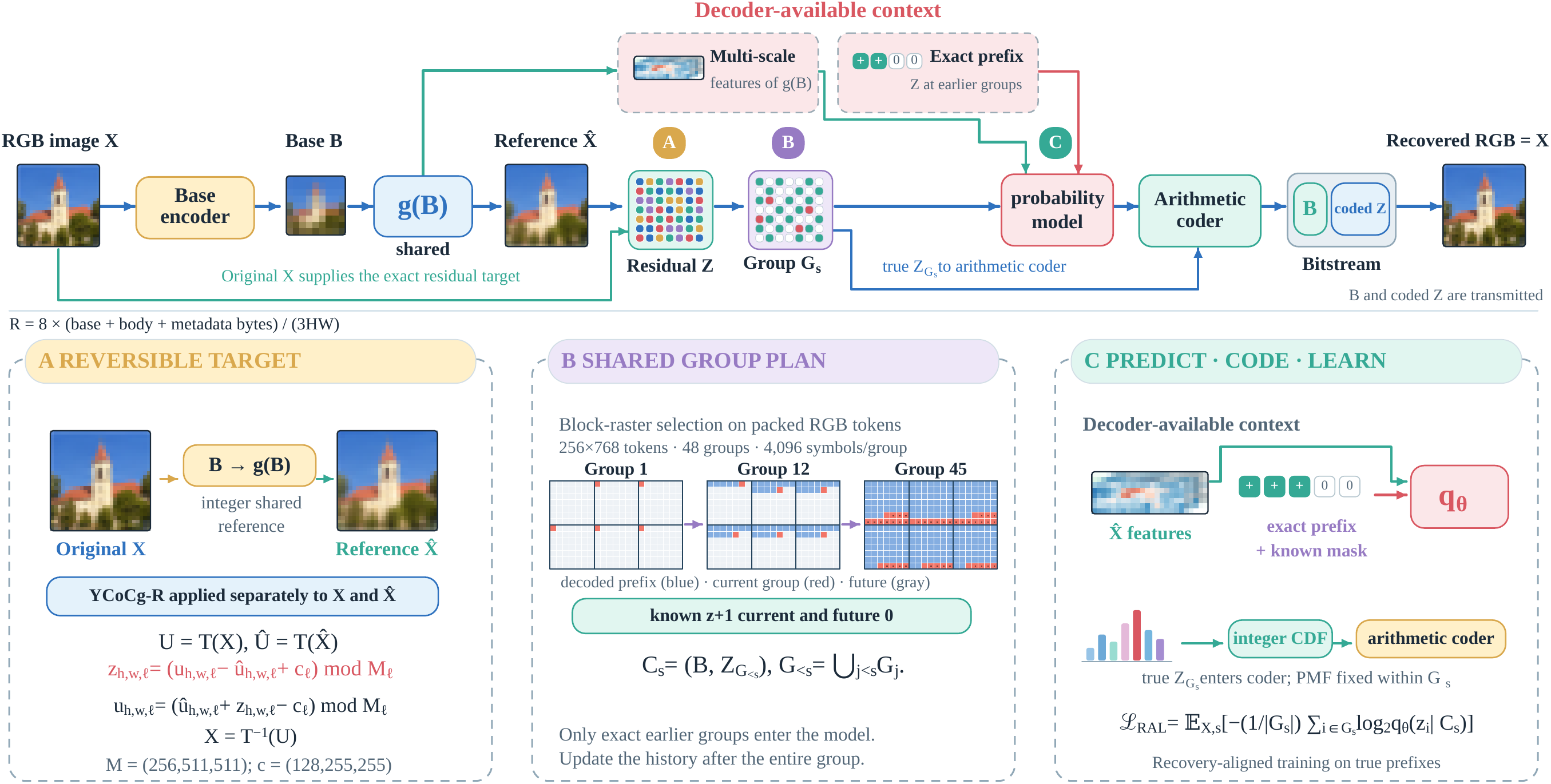}
    \caption{Codec architecture: end-to-end coding and exact reconstruction (top), with (A) reversible base-conditioned residual representation, (B) deterministic block-raster grouping, and (C) decoder-context-based prediction, arithmetic coding, and recovery-aligned training.}
    \label{fig:method-overview}
\end{figure*}

An overview of our codec is shown in Fig.~\ref{fig:method-overview}.
We first transmit a compact integer base and reconstruct a shared
reference at both the encoder and  decoder~\cite{mentzer2020lossy,bai2021joint,bai2024dlpr}.
The original image is then represented as reversible color residuals
relative to this reference. A deterministic block-raster plan
progressively partitions the residuals into groups. At each stage,
a neural probability model~\cite{larochelle2011nade,germain2015made,mentzer2018conditional,minnen2018joint}
predicts the current group from multi-scale features of the reconstructed
base and previously decoded residuals~\cite{oord2016pixelrnn,oord2016conditional,salimans2017pixelcnnpp,li2025masked},
and arithmetic coding encodes the resulting symbols
~\cite{rissanen1979arithmetic,witten1987arithmetic,howard1994arithmetic}.
We first formulate the resulting codelength, then describe the
residual representation, decoding schedule, and probability estimation.
Training and lossless decoding details are provided in the supplement.

\subsection{Problem Formulation and Codelength Decomposition}

Let $X\in\{0,\ldots,255\}^{H\times W\times3}$ be an RGB image, $N=3HW$, and $B$ the transmitted integer base. Both sides reconstruct the reference $\hat X=g(B)$. For each fixed $B$, the representation $Z=\Phi_B(X)$ is bijective on the valid image support~\cite{cover2006elements}, so
\begin{equation}
    H(Z\mid B)=H(X\mid B).
    \label{eq:conditional_entropy_invariance}
\end{equation}

Let $\pi=(G_0,\ldots,G_{S-1})$ be a deterministic ordered partition of the symbol positions, with $G_{<s}=\bigcup_{j<s}G_j$ and decoder context $C_s=(B,Z_{G_{<s}})$. Predicting symbols independently within each group~\cite{reed2017parallel} gives
\begin{equation}
    q_{\theta,\pi}(Z\mid B)
    =\prod_{s=0}^{S-1}\prod_{i\in G_s}q_\theta(Z_i\mid C_s).
    \label{eq:grouped_factorization}
\end{equation}
The expected ideal residual codelength~\cite{shannon1948communication,cover2006elements} decomposes as
\begin{equation}
    J=\mathbb E[-\log_2 q_{\theta,\pi}(Z\mid B)]
    =H(X\mid B)+\epsilon_{\mathrm{group}}+\epsilon_{\mathrm{fit}},
    \label{eq:codelength_decomposition}
\end{equation}
where
\begin{align}
    \epsilon_{\mathrm{group}}
    &=\sum_s\left[\sum_{i\in G_s}H(Z_i\mid C_s)
    -H(Z_{G_s}\mid C_s)\right],
    \label{eq:group_excess}
\end{align}
\begin{equation}
\varepsilon_{\mathrm{fit}}
=
\sum_s \sum_{i\in G_s}
\mathbb{E}_{C_s}
D_{\mathrm{KL}}\!\bigl(
p(Z_i\mid C_s)\|q(Z_i\mid C_s)
\bigr).
\label{eq:fit_excess}
\end{equation}
The nonnegative group term measures dependencies missed by independent within-group prediction~\cite{watanabe1960correlation}; the fit term measures mismatch to the true conditional marginals~\cite{kullback1951information}. An invertible color transform can change both terms by reorganizing dependencies. For fixed transformed coordinates and grouping, however, symbol-wise modular residualization conditioned on the same base preserves the group term; any coding benefit arises through model fit. Changing the grouping can affect both terms by changing the available history. The supplement gives the derivation. $J$ concerns the residual body; complete-file cost also includes the base and overhead.

\subsection{Reversible Base-Conditioned Residual Representation}

We apply the integer-reversible YCoCg-R transform~\cite{malvar2003ycocgr,malvar2008ycocg} $T$ separately to the image and its reference, obtaining $U=T(X)$ and $\hat U=T(\hat X)$. The Y, Co, and Cg lanes have sizes $(M_Y,M_{Co},M_{Cg})=(256,511,511)$, with chroma coordinates offset by 255. For pixel $(h,w)$ and lane $\ell$, the modular residual and its inverse are
\begin{align}
    z_{h,w,\ell}
    &=(u_{h,w,\ell}-\hat u_{h,w,\ell}+c_\ell)\bmod M_\ell,
    \label{eq:modular_residual}\\
    u_{h,w,\ell}
    &=(\hat u_{h,w,\ell}+z_{h,w,\ell}-c_\ell)\bmod M_\ell,
    \label{eq:inverse_modular_residual}
\end{align}
where $c=(128,255,255)$ and $X=T^{-1}(U)$. Transforming before subtraction matters because the integer lifting operations involve rounding~\cite{sweldens1996lifting,calderbank1998integers}.

For prediction, we interleave the lanes $\ell\in\{0,1,2\}$ along the spatial width:
\begin{equation}
    \tau_{h,3w+\ell}=z_{h,w,\ell}+1,
    \qquad \tau\in\{0,\ldots,511\}^{H\times3W}.
    \label{eq:packed_residual_tokens}
\end{equation}
Token $0$ denotes an unknown position; the valid tokens are $1$--$256$ for Y and $1$--$511$ for each chroma lane.

\subsection{Deterministic Block-Raster Conditional Factorization}

We partition the packed token plane into cells and select the first unknown position in each nonempty cell at every stage. If there are more than $C$ candidates, we retain approximately evenly spaced candidates; if fewer, we fill the group with the earliest remaining unknown positions in row-major order. The selected positions are sorted in row-major order. When at most $C$ positions remain, they form the final group. For a $256\times256$ tile, $C=4096$ gives 48 groups and a $37\times111$ cell grid. The supplement specifies the cell construction and selection rules.

This geometry-based plan yields spatially distributed groups and can be reproduced at both ends without transmitting position indices. The model predicts each group from $C_s$ in Eq.~\ref{eq:grouped_factorization}; its probabilities remain fixed until the whole group is recovered, after which the decoded symbols enter the history. Thus, earlier groups provide exact context while positions within a group are predicted in parallel~\cite{minnen2020channel,he2021checkerboard,he2022elic,kang2022pilc}.

\subsection{Base-Conditioned Probability Estimation}

We use a four-scale C256 SongUNet~\cite{song2021sde,karras2022edm,ronneberger2015unet} with multi-scale features from the reconstructed base. At stage $s$, the input contains the recovered token history $h^{(s)}$ (zero elsewhere), its known-position mask $m^{(s)}$, the known proportion $\rho_s=|G_{<s}|/N$, lane identities $L$, and the packed transformed reference $\hat\tau$:
\begin{equation}
\begin{aligned}
    A^{(s)}=\operatorname{Concat}\big[
    &\operatorname{Emb}_{24}(h^{(s)}),h^{(s)}/511,m^{(s)},\\
    &\rho_s\mathbf1,(1-\rho_s)\mathbf1,L,\hat\tau/511\big].
\end{aligned}
\label{eq:probability_input}
\end{equation}
The adapter has 32 channels. A four-level, 64-channel feature pyramid computed from $\hat\tau/511$ is projected into the corresponding encoder and decoder stages. The backbone uses a fixed time input $t_0=10^{-4}$ and outputs 512 logits per position; invalid categories are masked before normalization to obtain the lane-specific PMFs used for entropy coding.

\section{Experiments}
\label{sec:experiments}

Our experiments are organized around the two mechanisms investigated in this work: representation-induced predictability and decoder-available conditional structure. Since the experiments use different checkpoints, training protocols, and base-coding configurations, we report their results separately and avoid direct comparisons across unmatched settings.

\subsection{Experimental Protocol}
\label{sec:protocol}

For the codec and training comparisons, \textbf{BR-Path} uses block-raster (BR) recovery prefixes, while \textbf{Prior-Mask} uses prior-guided selection and masking. Both condition YCoCg-R residual prediction on a neural base and its multi-scale features. BR-Path uses a deterministic block-raster plan with 4,096-symbol groups, exact BR prefixes, selected-group supervision, and four sampled stage tasks per image.

We use three distinct evaluation settings. (1) Controlled representation and training ablations are conducted on 256 development images at $256\times256$ resolution, using a fixed $64\times64\times3$ base and 48 BR groups of 4,096 symbols. (2) The expanded inference-order diagnostic evaluates the frozen BR-Path and Prior-Mask checkpoints on the same 256-image evaluation set at $256\times256$ resolution. (3) Native-resolution evaluation uses the BR-Path codec with encoded base bitstream  on Kodak~\cite{kodakdataset}, DIV2K validation~\cite{agustsson2017div2k}, and CLIC2020 Professional and Mobile~\cite{clic2020dataset}, comprising 226 original-resolution images without resizing or cropping.

We measure predictive difficulty using conditional NLL under the actual decoded prefix, and evaluate compression efficiency using complete-file rates including the base payload, residual bitstream, and coding overhead. All evaluated bitstreams are verified by exact RGB reconstruction.

\subsection{Representation-Induced Predictability}
\label{sec:exp-repr}

We first ask whether information-preserving changes to the target representation alter the prediction difficulty of a practical neural probability model. We compare RGB and integer-reversible YCoCg-R coordinates, each with direct-symbol and base-residual prediction. All four configurations use the same base pipeline, explicit base features and BR decoding schedule. Thus, the color-space comparison changes the target and history coordinates rather than the entire base-coding pipeline.

\paragraph{Joint effects of color coordinates and residualization.}
Table~\ref{tab:representation} compares two complementary
representation choices: reversible color transformation
(RGB vs. YCoCg-R) and base-conditioned residualization
(direct vs. residual prediction). Both are evaluated under
the same base pipeline, grouping schedule, and training
protocol, allowing us to examine their individual and
combined effects on complete-file coding efficiency.

Residualization consistently reduces the coding rate in
both color spaces, but its benefit differs substantially:
0.2109 bpsp in RGB compared with 0.0766 bpsp in YCoCg-R.
Conversely, YCoCg-R reduces the rate by 0.1668 bpsp for
direct prediction, but only 0.0325 bpsp after
residualization. Thus, the advantage of YCoCg-R becomes
considerably smaller when prediction is performed on
base-conditioned residuals.

This non-additivity is further reflected in the
color--residual interaction of $-0.1343$ bpsp. Applying
both transformations yields a total reduction of
0.2434 bpsp relative to RGB direct prediction, substantially
smaller than the sum of their individual reductions
(0.3777 bpsp). The two representation choices therefore
provide overlapping rather than independent coding gains:
the benefit of changing color coordinates depends on
whether the model predicts direct symbols or base
residuals. This suggests that representation design should
be evaluated jointly rather than by optimizing each
transformation in isolation.

\begin{table}[!htbp]
\centering
\small
\caption{Target/history color coordinates and residual parameterization on the fixed 256-image development set.  The YCoCg-R direct and residual entries correspond to Direct + BR and Residual + BR in Table~\ref{tab:training-ablation}.}
\label{tab:representation}
\begin{tabular}{lrr}
\toprule
Color coordinates & Direct symbols & Base residual \\
\midrule
RGB & 2.793084 & 2.582182 \\
YCoCg-R & 2.626315 & \textbf{2.549699} \\
\bottomrule
\end{tabular}
\end{table}

\begin{figure*}[t]
\centering
\IfFileExists{fig/predictive_difficulty_4panel.pdf}{%
    \includegraphics[width=0.7\textwidth,height=0.5\textheight,keepaspectratio]
    {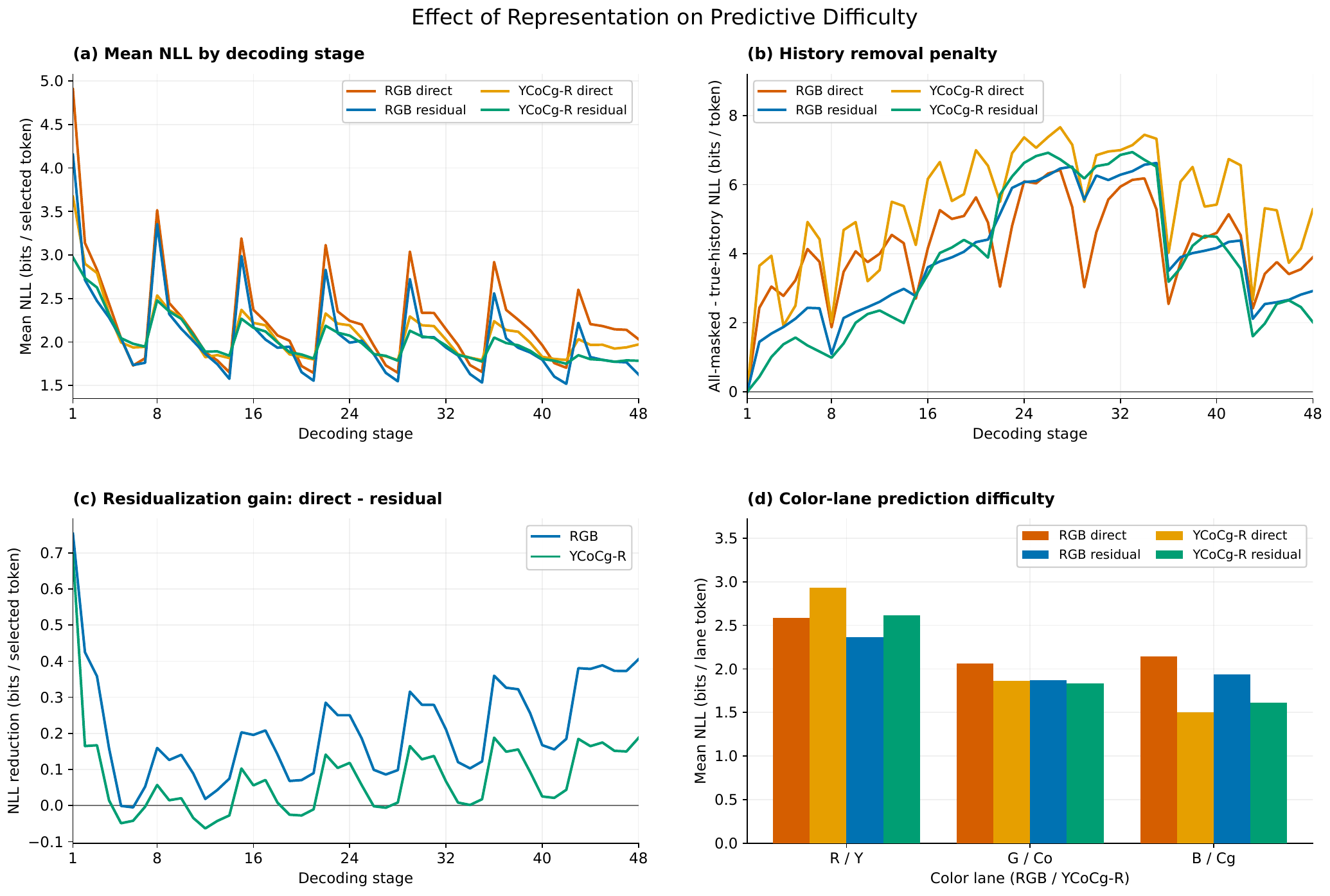}%
}{%
    \fbox{\parbox[c][2.4cm][c]{0.85\textwidth}
    {\centering Placeholder: representation-predictability figure}}%
}
\caption{Representation-induced predictive difficulty on the 256-image development set.
(a) Mean conditional NLL by decoding stage for the four target parameterizations.
(b) NLL penalty caused by removing previously decoded history.
(c) Stage-wise NLL gain from residualization (direct minus residual).
(d) Mean NLL across corresponding RGB and YCoCg-R color lanes.
NLL is evaluated using the true decoded prefix at each target's actual decoding stage, except for the history-removal intervention in (b).}
\label{fig:representation-predictability}
\end{figure*}

\paragraph{Stage-wise predictive difficulty.} We therefore measure the mean conditional NLL of the selected target tokens at each of the 48 decoding stages. Figure~\ref{fig:representation-predictability}(a) summarizes the stage-wise curves, while Fig.~\ref{fig:representation-predictability}(c) reports the direct--residual differences. The lane-wise NLLs are shown in Fig.~\ref{fig:representation-predictability}(d); the history intervention in panel (b) is discussed in Sec.~\ref{sec:exp-conditional}. In the early stages (1--12), YCoCg-R residuals have the lowest mean NLL (2.3073 bits per selected token), compared with 2.5951 for RGB direct, 2.3860 for YCoCg-R direct, and 2.4055 for RGB residuals. The same ordering holds over the middle stages (13--36), where YCoCg-R residuals achieve 1.9741 bits per selected token. The ordering changes in the late stages (37--48): RGB residuals achieve 1.8156 bits per selected token, slightly below the 1.8320 of YCoCg-R residuals.

This stage dependence is important: no single representation is uniformly easiest throughout the recovery process. At the complete-file level, residualization reduces the coding rate by 0.2109 bpsp in RGB and 0.0766 bpsp
in YCoCg-R, but its stage-wise gain is non-uniform. Some individual stages show a negative point-estimate gain, so the result should be interpreted as an average and stage-dependent effect rather than a uniform improvement at every stage.

RGB and YCoCg-R lanes are different coordinates, and the transform changes the symbol alphabets and the information represented by each lane. Taken together, the results show that reversible representation choices preserve the underlying information but alter the stage-wise conditional prediction difficulty and the distribution of model-achievable coding gains across coordinates.

\begin{table*}[!t]
\centering
\small
\setlength{\tabcolsep}{4pt}
\caption{Fixed-checkpoint inference-order diagnostic on the same 256 matched $256\times256$ images, in complete-file bpsp (lower is better). The five static plans use 4,096-symbol groups and 48 stages. The Prior-Mask checkpoint's softprior column reports its original 9,728-symbol, 21-stage configuration and is not directly matched to the other columns. All rates include the raw base and container.}
\label{tab:order-diagnostic}
\begin{tabular}{lrrrrrrr}
\toprule
Training recipe  & Block-raster & Raster & Pixel-major & Lane-major & Reverse & Softprior \\
\midrule
BR-Path  & \textbf{2.5907} & 4.6209 & 2.9584 & 4.9997 & 5.8271 & -- \\
Prior-Mask  & \textbf{2.6284} & 3.3582 & 2.6611 & 2.7198 & 2.6556 & 2.9467 \\
\bottomrule
\end{tabular}
\end{table*}

\subsection{Decoder-Available Conditional Structure}
\label{sec:exp-conditional}

The preceding analysis examines representation-induced
predictability under a fixed BR schedule. We now turn to
the complementary role of decoder-available conditional
information. We investigate this question at three levels:
the effect of decoding organization on coding efficiency,
the sensitivity of prediction to the content of decoded
history, and the relative contributions of exact history
and base-derived features.

\paragraph{Sensitivity to decoding-induced conditional structure.}
Table~\ref{tab:order-diagnostic} compares five decoding
plans using two frozen probability-model checkpoints.
Block-raster achieves the lowest complete-file rate for
both models, but the magnitude of its advantage differs
substantially. Relative to BR, alternative plans increase
the rate by 14.19\%--124.92\% for BR-Path, whereas the
corresponding increases for Prior-Mask range from only
1.03\% to 27.77\%.

This contrast reveals markedly different sensitivities
to the induced conditional structure. BR-Path exhibits
a pronounced preference for block-raster: even
pixel-major decoding incurs a noticeable rate increase,
while raster, lane-major, and reverse decoding lead to
substantially larger penalties. In contrast, Prior-Mask
maintains comparable rates under pixel-major,
lane-major, and reverse decoding, with raster remaining
the least favorable alternative. Thus, although
block-raster provides the best factorization for both
checkpoints, their predictive performance responds
differently to changes in group membership and available
history. This suggests that coding efficiency depends
not simply on the amount of accessible context, but on
how the induced conditional structure aligns with the
statistics learned by the probability model.

We further examine this effect by replacing the
prior-guided selector with block-raster under the
same 4,096-symbol grouping and 48-stage schedule for
Prior-Mask. BR reduces the complete-file rate from
2.8300 to 2.6284 bpsp, improving all 256 images.
This consistent gain indicates that the decoding
organization itself plays an important role in
determining the effectiveness of conditional prediction.

\paragraph{Sensitivity to decoded-history perturbations.}
We examine how the availability and fidelity of decoded
history affect prediction by comparing two interventions:
masking all previously decoded values and replacing them
with base-imputed symbols while retaining the known-position
mask. Figure~\ref{fig:representation-predictability}(b)
shows the all-masked penalty, while
Table~\ref{tab:history-intervention} reports both
interventions relative to the true-history baseline.

A consistent pattern emerges across all four
representations: replacing history with base-imputed
values incurs substantially larger NLL penalties than
removing it entirely. This indicates that the predictive
value of decoded history depends not only on its
availability but also on the fidelity of its content.
Moreover, the sensitivity ranking changes with the
intervention. Under all-masked history, YCoCg-R residuals
exhibit the smallest penalty, whereas under base-imputed
history, RGB residuals exhibit the largest. Residualization
also has opposite effects across the two interventions:
it reduces sensitivity to missing history but increases
sensitivity to base-imputed history in both spaces.
These results reveal that representation choices shape
not only how much the model benefits from decoded history,
but also how it responds to different forms of
history perturbation.

\begin{table}[!htbp]
\centering
\small
\setlength{\tabcolsep}{3pt}
\caption{History-intervention penalty, measured as intervention NLL minus NLL under the true decoded history. Values are mean bits/token over the common probe stages 4, 12, 24, 36, and 48.}
\label{tab:history-intervention}
\begin{tabular}{lrr}
\toprule
Representation & All-masked & Base-imputed \\
\midrule
RGB direct & 3.8607 & 15.6406 \\
YCoCg-R direct & 4.4212 & 7.3042 \\
RGB residual & 3.3938 & 18.0321 \\
YCoCg-R residual & 3.1102 & 10.1189 \\
\bottomrule
\end{tabular}
\end{table}

\paragraph{Relative roles of exact history and base-derived context.}
Following the history perturbation analysis, we further
examine the contributions of the two decoder-available
conditioning sources to actual coding efficiency.
Table~\ref{tab:context-ablation} compares the complete
model with variants that remove either exact decoded
values or the explicit base/FPN conditioning branch.

Removing exact history while retaining the known-position
mask increases the complete-file rate by 0.8357 bpsp.
This substantial increase shows that knowing which
symbols have been recovered is insufficient: their
actual values provide essential image-specific context
for predicting subsequent groups. Separately, removing the explicit base/FPN branch while retaining exact history increases the rate by 0.1816 bpsp relative to the full model, indicating that
multi-scale features from the reconstructed base provide
additional predictive information beyond the decoded
residual history. The larger penalty from removing exact
history highlights its dominant role in the current
configuration, while the base-derived features offer
complementary spatial context. Together, these results
show that effective conditional prediction relies on
both the content of previously recovered symbols and
the structural information provided by the shared base.

\paragraph{Implications for conditional prediction.}
Together, these experiments reveal that decoder-available
context affects coding efficiency through both its
organization and its content. The decoding schedule
determines which symbols are exposed as conditioning
information, while the predictive value of this history
depends on the fidelity of its decoded values. Exact
residual history and base-derived features further provide
complementary sources of context, with the former playing
a more prominent role in the evaluated configuration.
These findings indicate that effective conditional
factorization requires not only exposing useful dependencies
during decoding, but also providing the probability model
with informative and reliable conditioning signals.
\begin{table}[!htbp]
\centering
\small
\caption{Conditioning ablation under the residual and BR protocol. The base is transmitted in every configuration.}
\label{tab:context-ablation}
\setlength{\tabcolsep}{3pt}
\begin{tabular}{lccr}
\toprule
Config. & Base / FPN & History values & bpsp \\
\midrule
Residual + BR & On / On & On & \textbf{2.549699} \\
No History & On / On & Off & 3.385444 \\
No Base Features & Off / Off & On & 2.731274 \\
\bottomrule
\end{tabular}
\end{table}

\subsection{Effect of Recovery-Aligned Training}
\label{sec:exp-train}

We examine whether aligning the training objective with
the actual recovery process improves coding efficiency.
Table~\ref{tab:training-ablation} compares random-prefix
training with BR-prefix training under direct and
residual prediction. The BR-aligned recipe consistently
reduces the complete-file rate, with a substantially
larger gain for direct prediction than for residual
prediction (0.2716 vs. 0.0217 bpsp). Conversely,
residualization is beneficial under both training
recipes, but its gain is also reduced under BR-aligned
training (0.3265 vs. 0.0766 bpsp).

These results indicate that the benefit of recovery-aligned
training depends on the prediction target rather than
acting as an independent improvement. Aligning training
prefixes and supervised targets with the decoding process
provides a more effective training recipe for the evaluated
codec, while its contribution varies with the target
representation. 

\begin{table}[!htbp]
\centering
\small
\setlength{\tabcolsep}{4pt}
\caption{Training-recipe ablation on the 256-image development
set. All unrecovered and Next BR group denote supervision
over all positions outside the current prefix and only the next
block-raster group, respectively. Rates include the raw base
and container.}
\label{tab:training-ablation}
\begin{tabular}{llr}
\toprule
Configuration & Supervision & bpsp \\
\midrule
Direct + Random & All unrecovered & 2.897876 \\
Direct + BR & Next BR group & 2.626315 \\
Residual + Random & All unrecovered & 2.571400 \\
Residual + BR & Next BR group & \textbf{2.549699} \\
\bottomrule
\end{tabular}
\end{table}

\subsection{End-to-End Lossless Compression Performance}
\label{sec:exp-overall}

Finally, we evaluate the complete codec on 226
original-resolution images, accounting for the base
payload, residual bitstream, and container overhead.
Table~\ref{tab:overall} compares our method with
established lossless codecs and learned approaches.

Our method achieves lower rates than JPEG XL at
effort 11 on DIV2K and both CLIC subsets, while
remaining less efficient on Kodak. It also outperforms
the ArIB-BPS on these three
datasets. DLPR achieves lower rates across all four
datasets, indicating that a gap remains to the strongest
evaluated learned codec. Overall, these results establish
competitive complete-file coding performance across
multiple original-resolution image collections, while
highlighting the remaining room for improvement.

\begin{table}[!htbp]
\centering
\small
\setlength{\tabcolsep}{2.5pt}
\caption{Complete-file bpsp on original-resolution images. Lower is better. Best results are in bold and second-best results are underlined. Columns report Kodak, DIV2K, CLIC-p, and CLIC-m.}
\label{tab:overall}
\begin{tabular}{lrrrr}
\toprule
Method & Kodak & DIV2K & CLIC-p & CLIC-m \\
\midrule
PNG~\cite{w3c2003png}
& 4.4772 & 4.2338 & 3.8912 & 3.8783 \\

JPEG2000~\cite{itut2002jpeg2000}
& 3.1921 & 3.1180 & 2.9317 & 2.7073 \\

JPEG XL e9~\cite{alakuijala2019jpegxl}
& 2.8850 & 2.7871 & 2.6263 & 2.3589 \\

JPEG XL e11
& \textbf{2.7795} & 2.7201 & 2.5656 & 2.3122 \\

DLPR~\cite{bai2024dlpr}
& 2.8788 & \textbf{2.5394} & \textbf{2.3753} & \textbf{2.1648} \\

ArIB-BPS~\cite{zhang2024bps}
& \underline{2.7796} & 2.7709 & 2.6940 & 2.3051 \\

Ours
& 2.9688 & \underline{2.6809} & \underline{2.4884} & \underline{2.2367} \\
\bottomrule
\end{tabular}
\end{table}

\section{Conclusion}
\label{sec:conclusion}

We studied neural lossless image compression through the interplay of representation-induced predictability, decoder-available conditional structure, and recovery-aligned training. Our analysis separates the invariant conditional entropy preserved by reversible representations from the excess codelength arising from grouped prediction and imperfect model fitting.

Experiments reveal that representation choices provide non-additive, stage-dependent coding gains: the benefit of residualization depends on the color representation and decoding stage. Decoding organization also changes prediction difficulty and coding efficiency, with its effectiveness depending on the learned conditional structure. Exact decoded history provides the dominant conditioning signal, while base features offer complementary information. Recovery-aligned training further improves coding efficiency, with gains depending on the prediction target. On original-resolution benchmarks, our codec outperforms JPEG XL e11 on DIV2K and both CLIC subsets, but remains behind DLPR across all four datasets.

\section*{LLM Usage Statement}

ChatGPT (OpenAI) was used in this work for three purposes:
(1) assisting with English-language editing of the manuscript,
including grammar correction, wording refinement, and improvements
in clarity; (2) assisting with writing portions of the research
code; and (3) assisting with the visual refinement of the
Overview and Model Architecture figures, including layout
adjustments, formatting, and improvements in presentation clarity.
The authors take full responsibility for the technical content,
code correctness, figures, experimental results, and conclusions
presented in this paper.

{
  \small
  \bibliographystyle{ieeenat_fullname}
  \bibliography{main}
}
\onecolumn
\appendix
\numberwithin{equation}{section}
\numberwithin{figure}{section}
\numberwithin{table}{section}
\numberwithin{algorithm}{section}
\renewcommand{\theHequation}{\theHsection.\arabic{equation}}
\renewcommand{\theHfigure}{\theHsection.\arabic{figure}}
\renewcommand{\theHtable}{\theHsection.\arabic{table}}
\providecommand{\theHalgorithm}{\arabic{algorithm}}
\renewcommand{\theHalgorithm}{\theHsection.\arabic{algorithm}}

\setlength{\parindent}{1em}
\setlength{\parskip}{2pt}
\setlength{\emergencystretch}{1.2em}
\setlist{nosep,leftmargin=*}
\setlength{\textfloatsep}{8pt plus 2pt minus 1pt}
\setlength{\floatsep}{7pt plus 2pt minus 1pt}
\setlength{\intextsep}{7pt plus 2pt minus 1pt}
\raggedbottom
\renewcommand{\arraystretch}{1.15}

\begin{center}
  {\Large\bfseries Supplementary Material\par}
\end{center}

This supplementary material provides the theoretical justification, method definitions, experimental boundaries, and additional analyses that support the claims of the main paper. Section~\ref{sec:theory} proves the codelength decomposition and links its invariance properties directly to what each main-paper ablation isolates. Section~\ref{sec:recovery} specifies the block-raster factorization and recovery-aligned training objectives. Section~\ref{sec:protocols} defines the experimental protocols, data splits, and checkpoint correspondence across comparisons. Section~\ref{sec:diagnostics} presents the history-intervention definitions, matched-selector diagnostic, base-coding decomposition, and error-regime analysis, and Section~\ref{sec:impl} gives a concise architectural and lossless-coding overview. All entropies, KL divergences, and log-losses use base-two logarithms.

\section{Theoretical Proof and Experimental Implications}\label{sec:theory}

\subsection{Codelength Decomposition and Bijective Entropy Invariance}\label{sec:proof}

Let $X\in\{0,\ldots,255\}^{H\times W\times 3}$ be an RGB image, $B$ the transmitted integer base, and $Z=\Phi_B(X)$ a discrete representation that is bijective on the valid image support for every fixed $B$. Let $\pi=(G_0,\ldots,G_{S-1})$ be a deterministic ordered partition of the $N=3HW$ symbol positions, with prefix $G_{<s}=\bigcup_{j<s}G_j$ and decoder context $C_s=(B,Z_{G_{<s}})$. Predicting symbols conditionally independently within each group yields
\begin{equation}
  q_{\theta,\pi}(Z\mid B)=\prod_{s=0}^{S-1}\prod_{i\in G_s}q_\theta(Z_i\mid C_s),
  \qquad
  J(\theta,\Phi,\pi)=\E\!\left[-\log_2 q_{\theta,\pi}(Z\mid B)\right].
  \label{eq:a-grouped}
\end{equation}
Expanding the logarithm and applying the conditional cross-entropy identity at each position $i\in G_s$ separates the marginal conditional entropies from the neural estimation error:
\begin{equation}
  J = \sum_{s=0}^{S-1}\sum_{i\in G_s}H(Z_i\mid C_s)
      + \underbrace{\sum_{s=0}^{S-1}\sum_{i\in G_s}
        \E_{C_s}\!\left[\KL\!\left(p(Z_i\mid C_s)\Vert q_\theta(Z_i\mid C_s)\right)\right]}_{\epsilon_{\mathrm{fit}}}.
  \label{eq:a-ce}
\end{equation}
Adding and subtracting the group joint entropy $H(Z_{G_s}\mid C_s)$ within each stage isolates the conditional total correlation $\TC(Z_{G_s}\mid C_s)=\sum_{i\in G_s}H(Z_i\mid C_s)-H(Z_{G_s}\mid C_s)=\E_{C_s}[\KL(p(Z_{G_s}\mid C_s)\Vert\prod_{i\in G_s}p(Z_i\mid C_s))]\ge 0$:
\begin{equation}
  \sum_{s=0}^{S-1}\sum_{i\in G_s}H(Z_i\mid C_s)
  = \sum_{s=0}^{S-1}H(Z_{G_s}\mid C_s) + \sum_{s=0}^{S-1}\TC(Z_{G_s}\mid C_s)
  = H(Z\mid B) + \epsilon_{\mathrm{group}},
  \label{eq:a-tc}
\end{equation}
where the second equality uses the chain rule over the ordered partition $\pi$. Finally, because $\Phi_b$ is a discrete bijection for every realization $B=b$, it merely relabels outcomes without changing their probability masses, so $H(Z\mid B=b)=H(X\mid B=b)$ for every $b$ and hence $H(Z\mid B)=H(X\mid B)$. Combining Eqs.~\eqref{eq:a-ce} and~\eqref{eq:a-tc} gives the main-paper decomposition:
\begin{equation}
  J = H(X\mid B) + \epsilon_{\mathrm{group}}(Z,\pi) + \epsilon_{\mathrm{fit}}(Z,\pi,\theta),
  \label{eq:a-decomp}
\end{equation}
where $\epsilon_{\mathrm{group}}\ge 0$ quantifies within-group dependencies unexploitable under parallel prediction and $\epsilon_{\mathrm{fit}}\ge 0$ quantifies mismatch between the neural predictor and the true conditional marginals.

\subsection{Residualization Invariance and What Each Ablation Isolates}\label{sec:residual-corollary}

A crucial consequence of Eq.~\eqref{eq:a-decomp} is that different information-preserving operations act on different excess terms. Let $U=T(X)$ denote color-transformed image symbols and $\hat{U}=T(\hat{X})$ the transformed shared reference computed deterministically from $B$. Define the symbol-wise modular residual in lane $\ell(i)$ as
\begin{equation}
  Z_i = \psi_{B,i}(U_i) = \left(U_i - \hat{U}_i(B) + c_{\ell(i)}\right)\bmod M_{\ell(i)}.
  \label{eq:a-res-map}
\end{equation}
Conditioned on $B$, each $\hat{U}_i(B)$ is fixed and $\psi_{B,i}$ is a symbol-wise bijection on $\{0,\ldots,M_{\ell(i)}-1\}$. Therefore, the induced map from the prefix $U_{G_{<s}}$ to $Z_{G_{<s}}$ is also bijective, which implies for every stage $s$ and every position $i\in G_s$:
\begin{equation}
  H(Z_i\mid B,Z_{G_{<s}}) = H(U_i\mid B,U_{G_{<s}}),
  \qquad
  H(Z_{G_s}\mid B,Z_{G_{<s}}) = H(U_{G_s}\mid B,U_{G_{<s}}).
  \label{eq:a-prefix-inv}
\end{equation}
Consequently, $\TC(Z_{G_s}\mid C_s)=\TC(U_{G_s}\mid B,U_{G_{<s}})$ for every stage, so $\epsilon_{\mathrm{group}}(Z,\pi)=\epsilon_{\mathrm{group}}(U,\pi)$. Substituting this invariance into Eq.~\eqref{eq:a-decomp} yields a direct corollary:

\paragraph{Corollary (Pure model-fit effect of modular residualization).}
\emph{When the transmitted base $B$, color coordinates $T$, and deterministic grouping plan $\pi$ are held fixed, the ideal residual codelength difference between residual and direct symbol prediction equals purely the change in neural fitting error:}
\begin{equation}
  J_{\mathrm{res}} - J_{\mathrm{direct}}
  = \epsilon_{\mathrm{fit,res}} - \epsilon_{\mathrm{fit,direct}}.
  \label{eq:a-corollary}
\end{equation}

This identity sharpens the scientific interpretation of every controlled comparison in the main paper (summarized in Table~\ref{tab:a-mapping}). Because the controlled ablations hold the base payload and container format fixed, observed differences in complete-file rate directly reflect changes in residual codelength (up to minor arithmetic-coding overhead):
\begin{itemize}
  \item \textbf{YCoCg-R direct vs.\ residual (Main Table~1 and Table~5, \emph{Direct + BR} vs.\ \emph{Residual + BR}; also \emph{Direct + Random} vs.\ \emph{Residual + Random}):}
  Since the base, color coordinates, and grouping are identical, Eq.~\eqref{eq:a-corollary} applies directly. The observed $0.0766\;\bpsp$ gain (\emph{Direct + BR} $\to$ \emph{Residual + BR}) and $0.3265\;\bpsp$ gain (\emph{Direct + Random} $\to$ \emph{Residual + Random}) therefore appear unlikely to be explained by reductions in conditional entropy or within-group total correlation, and are instead consistent with a reduction in $\epsilon_{\mathrm{fit}}$.
  \item \textbf{RGB direct vs.\ residual (Main Table~1):} Provides a second controlled test of Eq.~\eqref{eq:a-corollary} in RGB coordinates. The much larger gain ($0.2109\;\bpsp$) demonstrates that the model-fit benefit of residualization depends strongly on the underlying coordinate system.
  \item \textbf{RGB vs.\ YCoCg-R (Main Table~1):} Unlike symbol-wise residualization, a cross-channel color transform mixes the three lanes of each pixel. While $H(Z\mid B)=H(X\mid B)$ remains invariant, cross-lane dependencies within and across groups are redistributed, allowing \emph{both} $\epsilon_{\mathrm{group}}$ and $\epsilon_{\mathrm{fit}}$ to change. The $-0.1343\;\bpsp$ interaction shows that color decorrelation and base residualization resolve overlapping sources of prediction difficulty.
  \item \textbf{Decoding-order changes (Main Table~2):} Changing $\pi$ alters both group memberships $G_s$ and the decoded histories $Z_{G_{<s}}$. Even under fixed representation and base, this can alter both $\epsilon_{\mathrm{group}}$ and $\epsilon_{\mathrm{fit}}$ (and in frozen-checkpoint tests, reflects alignment with the conditional statistics learned by the model).
\end{itemize}

\begin{table}[tbp]
\centering
\caption{Connection between the theoretical decomposition in Eq.~\eqref{eq:a-decomp} and the main-paper experiments. Every controlled comparison holds $H(X\mid B)$ invariant.}
\label{tab:a-mapping}
\small
\begin{tabularx}{\linewidth}{lYYc}
\toprule
Main-paper comparison & Fixed factors & Changed factor & Terms allowed to change\\
\midrule
Direct vs.\ Residual in YCoCg-R (Tables~1, 5) & Base, YCoCg-R coords, $\pi$, recipe & Symbol-wise modular residualization & $\epsilon_{\mathrm{fit}}$ only (Eq.~\eqref{eq:a-corollary})\\
Direct vs.\ Residual in RGB (Table~1) & Base, RGB coords, $\pi$, recipe & Symbol-wise modular residualization & $\epsilon_{\mathrm{fit}}$ only (Eq.~\eqref{eq:a-corollary})\\
RGB vs.\ YCoCg-R (Table~1) & Base pipeline, $\pi$, recipe & Cross-channel reversible bijection $T$ & $\epsilon_{\mathrm{group}}$ and $\epsilon_{\mathrm{fit}}$\\
Decoding orders (Table~2) & Base, representation, frozen $\theta$ & Partition $\pi$ (groups and prefixes) & $\epsilon_{\mathrm{group}}$ and $\epsilon_{\mathrm{fit}}$\\
Conditioning / Training recipe (Tables~4, 5) & Base, representation, inference $\pi$ & Conditioning inputs or training objective & $\epsilon_{\mathrm{fit}}$ only\\
\bottomrule
\end{tabularx}
\end{table}

\section{Conditional Recovery and Training Objectives}\label{sec:recovery}

\subsection{Deterministic Block-Raster Conditional Factorization}\label{sec:br}

For a packed token plane of size $H\times W_t$ ($W_t=3W$) and target group size $C$, the block-raster plan partitions the plane into a regular cell grid of dimensions
\begin{equation}
  n_h = \operatorname{round}\!\left(\sqrt{\frac{CH}{W_t}}\right),
  \qquad
  n_w = \left\lceil\frac{C}{n_h}\right\rceil,
  \label{eq:b-grid}
\end{equation}
mapping token coordinate $(r,c)$ to cell $(\lfloor r n_h/H\rfloor, \lfloor c n_w/W_t\rfloor)$. On a standard $256\times 256$ tile ($256\times 768$ tokens, $C=4{,}096$), this yields a $37\times 111$ grid ($4{,}107$ cells) and exactly $S=48$ groups of $4{,}096$ tokens.

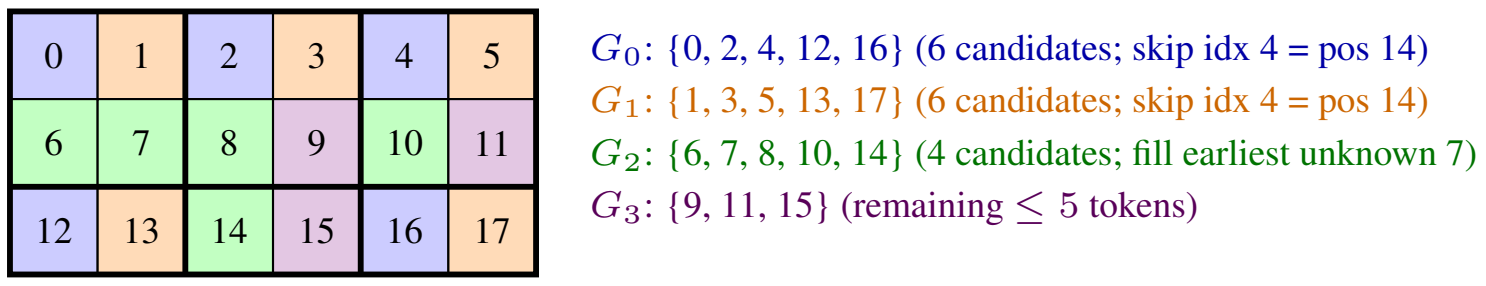
\begin{figure}[H]
\centering
\begin{tikzpicture}[x=0.58cm,y=0.58cm, every node/.style={font=\scriptsize}, line width=0.4pt]
\foreach \r/\c/\idx/\grp in {
  0/0/0/0,0/1/1/1,0/2/2/0,0/3/3/1,0/4/4/0,0/5/5/1,
  1/0/6/2,1/1/7/2,1/2/8/2,1/3/9/3,1/4/10/2,1/5/11/3,
  2/0/12/0,2/1/13/1,2/2/14/2,2/3/15/3,2/4/16/0,2/5/17/1} {
  \ifcase\grp\def\cellcolor{blue!20}\or\def\cellcolor{orange!28}\or\def\cellcolor{green!24}\or\def\cellcolor{violet!22}\fi
  \fill[\cellcolor] (\c,-\r) rectangle ++(1,-1);
  \draw (\c,-\r) rectangle ++(1,-1);
  \node at (\c+0.5,-\r-0.5) {\idx};
}
\draw[line width=1.1pt] (0,0) rectangle (6,-3);
\draw[line width=1.1pt] (2,0)--(2,-3);
\draw[line width=1.1pt] (4,0)--(4,-3);
\draw[line width=1.1pt] (0,-2)--(6,-2);
\node[anchor=west] at (6.4,-0.45) {\color{blue!65!black}$G_0$: \{0, 2, 4, 12, 16\} (6 candidates; skip idx 4 = pos 14)};
\node[anchor=west] at (6.4,-1.05) {\color{orange!80!black}$G_1$: \{1, 3, 5, 13, 17\} (6 candidates; skip idx 4 = pos 14)};
\node[anchor=west] at (6.4,-1.65) {\color{green!45!black}$G_2$: \{6, 7, 8, 10, 14\} (4 candidates; fill earliest unknown 7)};
\node[anchor=west] at (6.4,-2.25) {\color{violet!70!black}$G_3$: \{9, 11, 15\} (remaining $\le 5$ tokens)};
\end{tikzpicture}
\caption{Worked illustration of Algorithm~\ref{alg:b-br} on a $3\times 6$ token plane ($C=5$, $2\times 3$ cell grid). Thick lines delimit cells; colors indicate assigned groups. Skipped candidates remain eligible in later stages, and fill-in can draw multiple tokens from one cell once fewer than $C$ nonempty cells remain.}
\label{fig:b-3x6}
\end{figure}

Algorithm~\ref{alg:b-br} specifies the deterministic group construction. At each stage, every nonempty cell proposes its earliest currently unknown position in global row-major order. If more than $C$ candidates are proposed, $C$ indices are retained via evenly spaced integer truncation over the ordered candidate list (any skipped candidate remains unknown and re-enters eligibility at later stages); if fewer than $C$ candidates exist, the group is filled with the earliest remaining unknown positions in row-major order. Because the schedule depends solely on tile geometry, both encoder and decoder construct identical groups without transmitting position indices, while maintaining spatially spread context at every stage. Figure~\ref{fig:b-3x6} illustrates the mechanism on a $3\times 6$ token plane with $C=5$ and a $2\times 3$ cell grid.

\begin{algorithm}[H]
\caption{Deterministic cell-based block-raster grouping}
\label{alg:b-br}
\small
\begin{suppalgorithmic}
\AlgRequire Packed dimensions $H,W_t$, target group size $C\ge 1$, valid cell grid $(n_h,n_w)$
\AlgEnsure Ordered partition $\pi=(G_0,\ldots,G_{S-1})$
\AlgState Initialize unknown set $\mathcal{U}\gets\{0,\ldots,HW_t-1\}$ in global row-major order; precompute cell IDs
\AlgState Initialize empty schedule $\pi$
\AlgWhile{$|\mathcal{U}|>0$}
  \AlgIf{$|\mathcal{U}|\le C$}
    \AlgState $G \gets \mathcal{U}$
  \AlgElse
    \AlgState Collect the earliest position in $\mathcal{U}$ from each nonempty cell to form row-major sorted list $\mathcal{C}$
    \AlgIf{$|\mathcal{C}|>C$}
      \AlgState Retain $C$ evenly spaced integer indices from $\{0,\ldots,|\mathcal{C}|-1\}$ to form $G$
    \AlgElse
      \AlgState $G \gets \mathcal{C}$; append earliest positions in $\mathcal{U}\setminus G$ until $|G|=C$
    \AlgEndIf
  \AlgEndIf
  \AlgState Sort $G$ in row-major order, append $G$ to $\pi$, and update $\mathcal{U}\gets\mathcal{U}\setminus G$
\AlgEndWhile
\AlgState \AlgReturn $\pi$
\end{suppalgorithmic}
\end{algorithm}

\FloatBarrier

\subsection{Decoder-Aligned Context and Training Objectives}\label{sec:objectives}

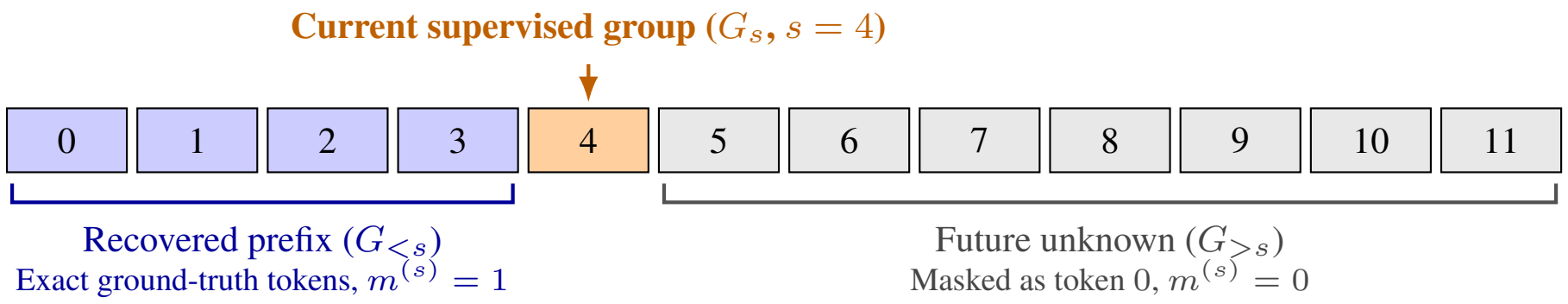
\begin{figure}[H]
\centering
\begin{tikzpicture}[x=1.0cm,y=0.72cm, every node/.style={font=\footnotesize}]
  \foreach \i in {0,...,11} {
    \ifnum\i<4
      \fill[blue!20] (\i,0) rectangle ++(0.92,0.68);
    \else\ifnum\i=4
      \fill[orange!38] (\i,0) rectangle ++(0.92,0.68);
    \else
      \fill[gray!18] (\i,0) rectangle ++(0.92,0.68);
    \fi\fi
    \draw (\i,0) rectangle ++(0.92,0.68);
    \node at (\i+0.46,0.34) {\i};
  }
  \draw[-{Latex[length=2mm]},thick,orange!75!black] (4.46,1.15) -- (4.46,0.74);
  \node[anchor=south,orange!75!black] at (4.46,1.18) {\textbf{Current supervised group ($G_s$, $s=4$)}};

  \draw[thick,blue!60!black] (0.04,-0.12) -- (0.04,-0.32) -- (3.88,-0.32) -- (3.88,-0.12);
  \node[anchor=north,blue!60!black,align=center] at (1.96,-0.40) {Recovered prefix ($G_{<s}$)\\[-1pt]\scriptsize Exact ground-truth tokens, $m^{(s)}=1$};

  \draw[thick,gray!65!black] (5.04,-0.12) -- (5.04,-0.32) -- (11.88,-0.32) -- (11.88,-0.12);
  \node[anchor=north,gray!55!black,align=center] at (8.46,-0.40) {Future unknown ($G_{>s}$)\\[-1pt]\scriptsize Masked as token $0$, $m^{(s)}=0$};
\end{tikzpicture}
\caption{Schematic of the recovery-aligned training context (12 stages shown for illustration; standard tiles use 48 stages). Exact tokens populate $G_{<s}$, current and future tokens are masked, and cross-entropy loss is accumulated exclusively over $G_s$.}
\label{fig:b-context}
\end{figure}

Let $\phi(i)\in\{0,\ldots,47\}$ denote the group index of position $i$ under Algorithm~\ref{alg:b-br}. For any stage $s$, the exact decoder-available history and mask are constructed directly from the ground-truth tokens $\tau_i$ without rolling out earlier stages:
\begin{equation}
  h_i^{(s)} = \tau_i\cdot\mathbb{I}[\phi(i)<s],
  \qquad
  m_i^{(s)} = \mathbb{I}[\phi(i)<s].
  \label{eq:b-context}
\end{equation}
Figure~\ref{fig:b-context} illustrates how the recovered prefix, current supervised group, and future masked tokens partition the schedule at stage $s$.

\paragraph{Stratified K4 sampling and unbiasedness.}
For each source-image presentation, the K4 recipe evaluates $K=4$ stratified stages spaced by $12$ around the 48-stage cycle:
\begin{equation}
  \mathcal{S}(r) = \left\{(r + 12k)\bmod 48 \;\middle|\; k\in\{0,1,2,3\}\right\},
  \label{eq:b-k4}
\end{equation}
where the starting offset $r\in\{0,\ldots,47\}$ has a uniform marginal distribution across $\{0,\ldots,47\}$ (and is paired across controlled ablation configurations so that all configurations observe identical image--stage sequences). Every stage $s\in\{0,\ldots,47\}$ therefore has marginal inclusion probability $1/12$. Accumulating cross-entropy solely on the selected group $G_s$ gives the recovery-aligned per-image objective
\begin{equation}
  \mathcal{L}_{\BR}(X,r)
  = -\frac{1}{KC}\sum_{s\in\mathcal{S}(r)}\sum_{i\in G_s}\log_2 q_\theta(z_i\mid C_s).
  \label{eq:b-ral}
\end{equation}
Because $(KC)^{-1}\cdot(1/12)=(48C)^{-1}$ for $K=4$, taking the expectation over $r$ yields
\begin{equation}
  \E_r\!\left[\mathcal{L}_{\BR}(X,r)\mid X\right]
  = -\frac{1}{48C}\sum_{s=0}^{47}\sum_{i\in G_s}\log_2 q_\theta(z_i\mid C_s),
  \label{eq:b-unbiased}
\end{equation}
which is an unbiased estimator of the full 48-stage ideal residual codelength per symbol for image $X$. The backbone still computes logits over the full token plane before indexing $G_s$, so selected-only supervision aligns the training gradient with decoding rather than reducing U-Net forward FLOPs.

\paragraph{Random-prefix baseline objective (Main Table~5, Direct/Residual + Random).}
To evaluate recovery-aligned training fairly, the random-prefix baselines sample the same stratified stage indices $s\in\mathcal{S}(r)$ but replace the block-raster prefix with a uniformly random known set $R_s$ of size $sC$ (sampled independently per stage context) and supervise all remaining $|U_s|=(48-s)C$ unknown tokens. Within each context, cross-entropy is normalized by $|U_s|$, and contexts are averaged with equal weight $1/K$:
\begin{equation}
  \mathcal{L}_{\mathrm{rand}}(X,r)
  = \frac{1}{K}\sum_{s\in\mathcal{S}(r)}
    \left(-\frac{1}{|U_s|}\sum_{j\in U_s}\log_2 q_\theta\!\left(v_j\mid B, v_{R_s}, R_s\right)\right),
  \label{eq:b-rand}
\end{equation}
where $v_j$ denotes direct or residual symbols. Because random-prefix training and BR-aligned training jointly differ in prefix geometry (random vs.\ block-raster) and supervised support (all unknown vs.\ next group, averaging $98C$ vs.\ $4C$ supervised token events per presentation), Main Table~5 evaluates complete training recipes rather than isolating prefix geometry from target selection.

\section{Experimental Protocols and Scope of Comparisons}\label{sec:protocols}

Because our paper investigates both controlled component mechanisms and end-to-end native-resolution compression, different experiments use distinct checkpoints, evaluation subsets, and base-storage configurations. Table~\ref{tab:c-protocols} maps every experiment to its exact protocol so that comparisons are restricted strictly to matched conditions.

\begin{table}[tbp]
\centering
\caption{Experimental protocols, model lineages, and scope of valid comparisons across the main paper and supplementary studies.}
\label{tab:c-protocols}
\small
\setlength{\tabcolsep}{3.5pt}
\renewcommand{\arraystretch}{1.12}
\begin{tabularx}{\linewidth}{>{\raggedright\arraybackslash}p{0.22\linewidth}YYY}
\toprule
Study & Model / training lineage & Evaluation data & Fixed coding protocol\\
\midrule
Controlled ablations (Main Tables~1, 3, 4, 5; Fig.~3) & Standardized 9k configurations (same init, seed~42, paired K4 data order, 9k updates) & 256-image development set ($256\times 256$ resized center crops) & Frozen base $g(B)$, raw base ($0.5\;\bpsp$), BR4096 / 48 stages\\
Frozen inference-order \& matched-selector diagnostics (Main Table~2; Sec.~\ref{sec:softprior}) & \texttt{BR-Path} ($+1$k steps on the 9k \emph{Residual + BR} trajectory, 10k updates) vs.\ frozen \texttt{Prior-Mask} (earlier multi-stage lineage, 23k updates at batch~64) & Separate 256-image benchmark subset ($256\times 256$ resized center crops; partial overlap with dev set) & Raw base; 5 static plans at 4096/48 (+ original Softprior 9728/21; matched Softprior 4096/48)\\
Base-storage ablation (Table~\ref{tab:d-nbec}) & Frozen \texttt{Prior-Mask} deployed under BR4096 & Separate 256-image reserve subset ($256\times 256$ resized center crops) & Raw vs.\ NBEC base; identical BR4096 residual stream\\
Native-resolution benchmark (Main Table~6) & 40k domain-adapted \texttt{BR-Path} (20k general pool + 20k native DIV2K/CLIC tiles) & 226 original-res images (Kodak24, DIV2K100, CLIC-p41, CLIC-m61) & NBEC base; edge-replicated $256\times 256$ tiles; BR4096 / 48 stages\\
\bottomrule
\end{tabularx}
\end{table}

\paragraph{Data preprocessing, splits, and deduplication.}
For the standardized 9k controlled ablations, training and development images are drawn from disjoint train and evaluation archives of an internal ImageNet synset JPEG pool ($768{,}000$ training records containing $744{,}662$ unique decoded RGB images; $256$ development images). Every image is decoded to RGB, bicubic-resized so the shorter side equals $256$, and center-cropped to $256\times 256$ (no random cropping or flipping). Exact audits confirm zero overlap between the training pool and the 256-image development set across source IDs, raw file hashes, and decoded RGB hashes. The three 256-image subsets used in the controlled ablations, frozen inference-order diagnostic, and base-storage ablation are distinct subsets drawn from the evaluation archive with only partial image overlap; hence rates are never compared across subsets.

\paragraph{Native-resolution domain adaptation.}
The 40k codec in Main Table~6 starts from a separate 20k-update \texttt{BR-Path} checkpoint pretrained on the resized ImageNet pool and continues for $20{,}000$ updates on $2{,}433$ native-resolution photographic training images (DIV2K train 800, CLIC2020 Professional train 585, CLIC2020 Mobile train 1,048), partitioned without resizing into $106{,}027$ fixed $256\times 256$ tiles with edge replication (tile shares: $35.8\%$ DIV2K, $22.2\%$ CLIC-p, $42.0\%$ CLIC-m) and shuffled at the tile level while keeping $g(B)$ and NBEC frozen. File-hash and decoded-RGB audits confirm zero overlap between these training sources and the 226 original-resolution test images.

\paragraph{Important protocol distinctions.}
First, in the frozen inference-order diagnostic (Main Table~2), \texttt{Prior-Mask} is an earlier multi-stage checkpoint trained with a different recipe ($K=2$, chunk size $9{,}728$, Softprior selection, $5\%$--$25\%$ extra reveals, 23k total updates at batch size 64, i.e., $1.47\text{M}$ presentations vs.\ $2.56\text{M}$ for 10k \texttt{BR-Path}); it therefore serves as an independent training-recipe comparator rather than an equal-budget match. Second, in the color-coordinate ablation (Main Table~1), all four models share the identical YCoCg-R base reconstruction and auxiliary base/FPN pipeline while changing only the target and history coordinates (RGB direct uses $\tau=X+1$; RGB residual uses $((X-\hat{X}+128)\bmod 256)+1$ with 256 valid categories per lane inside the shared 512-logit head). Third, in the conditioning ablation (Main Table~4), the \emph{No Base Features} configuration zeroes the base scalar and disables the FPN branch while still transmitting $B$ for inverse residualization, so it isolates the value of explicit base/FPN conditioning rather than the net cost of transmitting a base. Finally, in Main Table~6, the ArIB-BPS baseline uses the official ImageNet64-pretrained (\texttt{IN64}) weights evaluated on $256\times 256$ tiles with an explicit zero-state fallback for numerical stability, with all fallback bytes included in the complete-file rate.

\paragraph{Optimization summary.}
Table~\ref{tab:c-optim} lists the shared optimization settings used for all standardized 9k ablation configurations.

\begin{table}[tbp]
\centering
\caption{Training budget and optimization settings shared by all standardized 9k controlled ablation configurations.}
\label{tab:c-optim}
\small
\setlength{\tabcolsep}{5pt}
\begin{tabularx}{\linewidth}{>{\raggedright\arraybackslash}p{0.32\linewidth}Y}
\toprule
Setting & Value\\
\midrule
Optimizer & Adam ($\beta_1=0.9$, $\beta_2=0.999$, $\epsilon=10^{-8}$, weight decay $0$)\\
Learning rate \& schedule & $5\times 10^{-5}$; 500-step linear warmup, then constant\\
Gradient clipping / EMA & Global norm $1.0$ / None\\
Precision (training / coding) & \texttt{bf16} mixed precision (activation checkpointing) / \texttt{fp32}\\
Global batch size \& updates & 256 source images; 9,000 updates ($2.304\text{M}$ presentations; $9.216\text{M}$ K4 contexts)\\
Random seed / base network & Seed 42 (paired image--stage trace) / Frozen $g(B)$\\
\bottomrule
\end{tabularx}
\end{table}

\section{Additional Analysis and Diagnostics}\label{sec:diagnostics}

\subsection{History-Intervention Definitions and Interpretation}\label{sec:history-int}

Main Table~3 evaluates inference-time prefix perturbations on the four frozen 9k color-ablation models over the five non-trivial probe stages $\{4,12,24,36,48\}$ (one-based; excluding stage~1 where the prefix is empty). For representation $a$, both interventions hold the known-position mask $m^{(s)}$, global progress $\rho_s$, base conditioning, and true target symbols fixed while keeping current and future groups masked at $0$. Under \emph{all-masked}, every prefix token $i\in G_{<s}$ is set to $0$. Under \emph{base-imputed}, every prefix position $i\in G_{<s}$ is replaced by the representation-consistent token obtained when the source image equals the shared base reconstruction ($\hat{X}$ in place of $X$):
\begin{equation}
  \beta_i^{(a)}=
  \begin{cases}
    \hat{X}_i+1, & a=\mathrm{RGB\ direct},\\
    \hat{U}_i+1, & a=\mathrm{YCoCg\text{-}R\ direct},\\
    129, & a=\mathrm{RGB\ residual},\\
    c_{\ell(i)}+1\;\;(129\text{ for Y},\;256\text{ for Co/Cg}), & a=\mathrm{YCoCg\text{-}R\ residual}.
  \end{cases}
  \label{eq:d-impute}
\end{equation}

Because every $\beta_i^{(a)}>0$, the model receives a syntactically valid prefix indicating that positions in $G_{<s}$ have been recovered, but with zero-residual (base-only) content. Across all four representations (Main Table~3), base-imputed history incurs a far larger NLL penalty than masking the prefix altogether (e.g., $10.12$ vs.\ $3.11$ bits/token for YCoCg-R residuals, and $18.03$ vs.\ $3.39$ for RGB residuals). This asymmetry confirms two points: first, the predictor genuinely exploits fine-grained, image-specific values in $Z_{G_{<s}}$ rather than using the mask alone; second, residualized predictors become \emph{less} fragile to absent history (because the residual distribution remains centered around zero) yet \emph{more} sensitive to false zero-residual history (which signals an artificially flat residual field contradicting the true high-frequency content). Consistent with this diagnostic, retraining without exact history values (Main Table~4, \emph{No History}) degrades complete-file rate by $0.8357\;\bpsp$ (worse on 255/256 images), whereas removing the explicit base/FPN branch degrades rate by $0.1816\;\bpsp$ (worse on all 256/256 images).

\subsection{Softprior Selector and Matched-Schedule Comparison}\label{sec:softprior}

In Main Table~2, \texttt{Prior-Mask} refers to the frozen model checkpoint, whereas \texttt{Softprior} refers to its confidence-based decoding selector. At each stage, \texttt{Softprior} scores each unknown token $i$ by its maximum predicted probability $\max_v q_\theta(v\mid C_s)$, adds a soft luminance priority bonus $+0.035\cdot\min(1, 3f_Y)$ to unknown Y-lane tokens (where $f_Y$ is the remaining fraction of unknown Y tokens), and selects the top-$C$ positions (ties broken by row-major index). Because selection depends only on decoder-available probabilities, no position indices are transmitted.

In Main Table~2, the \texttt{Softprior} column reports \texttt{Prior-Mask}'s native $C=9{,}728$, 21-stage configuration ($2.9467\;\bpsp$), which has fewer stages than the 48-stage static schedules. To eliminate this stage-count mismatch, we ran a matched-selector comparison on the exact same frozen \texttt{Prior-Mask} checkpoint and 256-image benchmark set with both \texttt{Softprior} and \texttt{Block-raster} fixed to $C=4{,}096$ and 48 stages. Replacing \texttt{Softprior} with \texttt{Block-raster} reduces the complete-file rate from $2.8300$ to $2.6284\;\bpsp$, improving all 256/256 images. Thus, the spatially uniform context exposed by block-raster improves conditional prediction even for a model trained specifically under confidence-guided selection.

\subsection{Base-Coding Decomposition: Raw Storage vs.\ NBEC}\label{sec:nbec-ablation}

While the controlled ablations store $B$ as raw uint8 bytes ($0.5\;\bpsp$) to keep residual differences transparent, the native-resolution codec compresses $B$ losslessly using Native Base Entropy Coding (NBEC; Section~\ref{sec:impl-rep}). Table~\ref{tab:d-nbec} isolates the byte-level effect of NBEC on the 256-image reserve subset with frozen \texttt{Prior-Mask} under BR4096. Because NBEC reconstructs the exact same integer base $B$, the residual bitstream is completely unchanged ($2.026813\;\bpsp$). NBEC lowers the base cost by $0.235077\;\bpsp$ at a small container increase ($+0.013631\;\bpsp$), yielding a net complete-file reduction of $0.221446\;\bpsp$ ($8.67\%$, improving all 256/256 images). This confirms that base compression and residual probability modeling decouple cleanly; because Table~\ref{tab:d-nbec} uses a separate subset, its saving should not be subtracted directly from the raw-base ablation tables.

\begin{table}[tbp]
\centering
\caption{Byte-level decomposition comparing raw base storage and NBEC on the 256-image reserve subset (frozen \texttt{Prior-Mask} under BR4096/48, in $\bpsp$).}
\label{tab:d-nbec}
\small
\begin{tabular}{lcc}
\toprule
Bitstream component & Raw base & NBEC base\\
\midrule
Base payload & 0.500000 & 0.264923\\
Residual body (bit-identical) & 2.026813 & 2.026813\\
Container & 0.027832 & 0.041463\\
\midrule
Complete file & 2.554645 & 2.333199\\
\bottomrule
\end{tabular}
\end{table}

\subsection{Stage-Wise Behavior and Benchmark Regimes}\label{sec:regimes}

Two boundary behaviors help clarify the scope of our empirical findings:
\begin{itemize}
  \item \textbf{Stage-dependent crossover in predictive difficulty (Main Fig.~3):} YCoCg-R residuals achieve the lowest mean NLL in early stages $1$--$12$ ($2.3073$ bits/token vs.\ $2.4055$ for RGB residuals) and middle stages $13$--$36$ ($1.9741$ bits/token), where cross-channel decorrelation most strongly aids sparse-context prediction. In late stages $37$--$48$, once dense local neighbors have been recovered, RGB residuals slightly outperform YCoCg-R residuals ($1.8156$ vs.\ $1.8320$ bits/token). Representation gains are therefore average and stage-dependent rather than uniform at every stage.
  \item \textbf{High-resolution gains vs.\ Kodak24 gap (Main Table~6):} Our 40k codec beats JPEG XL e11 on DIV2K ($-1.44\%$), CLIC-p ($-3.01\%$), and CLIC-m ($-3.27\%$) as well as ArIB-BPS on all three ($-3.25\%$, $-7.63\%$, $-2.97\%$), but trails both on Kodak24 ($2.9688$ vs.\ $2.7795\;\bpsp$) and remains behind DLPR across all four benchmarks. Kodak24 images are scanned film photographs outside the DIV2K/CLIC fine-tuning domain, and our fixed independent $256\times 256$ tiling with a non-adaptive NBEC base table cannot exploit global image statistics across tiles as aggressively as specialized codecs like JPEG XL or DLPR.
\end{itemize}

\section{Implementation Overview}\label{sec:impl}

This section summarizes the reversible representation, architecture, and lossless coding pipeline needed to understand the system; low-level container layouts and numerical coder serialization details belong in the code documentation.

\subsection{Reversible Color-Residual Representation and Tiling}\label{sec:impl-rep}

\paragraph{Base construction, shared upsampler $g(B)$, and NBEC.}
For each $256\times 256$ uint8 RGB tile $X$, the encoder forms a $64\times 64\times 3$ integer base $B$ via area downsampling followed by clamping and rounding to uint8. Both encoder and decoder pass the transmitted integer tensor $B$ through a frozen neural upsampler $g(B)$ combining $\times 4$ bicubic interpolation of $B/255$ with a bounded correction $0.25\tanh(\Delta(B))$ predicted by a lightweight 4-residual-block CNN with $\text{PixelShuffle}(4)$, rounded and clamped to uint8 RGB $\hat{X}$ (pretrained once on the ImageNet pool with weighted YCoCg-R $L_1$ + RGB anchor loss). When NBEC is enabled, $B$ is losslessly encoded via reversible modular YCoCg-R, causal median prediction, modulo-256 residuals split into high/low 4-bit nibbles, and arithmetic coding using fixed context tables (21 lane--edge contexts fitted once on 4,096 training bases).

\paragraph{YCoCg-R lifting and modular residuals.}
All lifting steps use signed integers with floor division (writing the blue channel as $B_c$):
\begin{align}
  \text{Forward:}\quad & Co=R-B_c,\;\; t=B_c+\lfloor Co/2\rfloor,\;\; Cg=G-t,\;\; Y=t+\lfloor Cg/2\rfloor, \label{eq:f-fwd}\\
  \text{Inverse:}\quad & t=Y-\lfloor Cg/2\rfloor,\;\; G=Cg+t,\;\; B_c=t-\lfloor Co/2\rfloor,\;\; R=Co+B_c. \label{eq:f-inv}
\end{align}
Chroma channels are shifted by $+255$ to obtain nonnegative coordinates $U=(Y, Co+255, Cg+255)$ with moduli $(M_Y,M_{Co},M_{Cg})=(256,511,511)$ and centering offsets $(c_Y,c_{Co},c_{Cg})=(128,255,255)$. Applying the transform separately to $X$ and $\hat{X}$ before subtraction is required because integer floor division is nonlinear. The modular residual and inverse are
\begin{equation}
  z_{h,w,\ell} = (u_{h,w,\ell}-\hat{u}_{h,w,\ell}+c_\ell)\bmod M_\ell,
  \qquad
  u_{h,w,\ell} = (\hat{u}_{h,w,\ell}+z_{h,w,\ell}-c_\ell)\bmod M_\ell,
  \label{eq:f-mod}
\end{equation}
followed by subtracting $255$ from $u_{Co},u_{Cg}$ and applying Eq.~\eqref{eq:f-inv} to recover $X$ exactly. Residuals are shifted by $+1$ (reserving token $0$ for unknown positions) and lane-interleaved along width into $\tau_{h,3w+\ell}=z_{h,w,\ell}+1$, giving valid token IDs $1$--$256$ for Y and $1$--$511$ for Co/Cg.

\paragraph{Native-resolution tiling and rate accounting.}
Arbitrary-resolution images are partitioned from the top-left into non-overlapping $256\times 256$ tiles, extending right and bottom boundary tiles by \textbf{edge replication}. Each tile is coded independently with its own base and 48-stage residual stream (no cross-tile history sharing), decoded independently, and cropped back to $(H,W)$. Complete-file rate in bits per RGB scalar value ($\bpsp$, with $\bpp=3\times\bpsp$) is computed directly from verified encoded files: the numerator includes all base, residual, padding, and container bytes, while the denominator counts only the unpadded original $3HW$ RGB scalars. Shared model weights and static NBEC tables are not charged per image.

\subsection{Probability Backbone Summary}\label{sec:impl-backbone}

\begin{figure}[tbp]
\centering
\resizebox{\linewidth}{!}{%
\begin{tikzpicture}[
  node distance=0.8cm and 0.95cm,
  block/.style={draw,rounded corners,align=center,minimum height=0.95cm,minimum width=2.2cm,fill=blue!6,font=\small},
  input/.style={draw,rounded corners,align=center,minimum height=0.85cm,minimum width=2.3cm,fill=gray!12,font=\small},
  arrow/.style={-{Latex[length=2mm]},thick}
]
\node[input] (hist) {Residual history\\and known mask};
\node[input,below=of hist] (base) {Decoded base\\reference $\hat{X}=g(B)$};
\node[block,right=of hist] (adapter) {32-channel\\input adapter};
\node[block,right=of base] (fpn) {4-level base FPN\\(64 channels)};
\coordinate (midright) at ($(adapter.east)!0.5!(fpn.east)$);
\node[block,right=1.0cm of midright] (unet) {C256\\SongUNet};
\node[block,right=of unet] (head) {GN + SiLU\\$3\times 3$ Conv};
\node[block,right=of head] (pmf) {Lane mask\\+ Softmax};
\coordinate (tap) at ($(base.east)!0.42!(fpn.west)$);
\coordinate (corridor) at ($(adapter.south)!0.5!(fpn.north)$);
\draw[arrow] (hist.east) -- (adapter.west);
\draw[arrow] (base.east) -- (fpn.west);
\fill (tap) circle (1.6pt);
\draw[arrow] (tap) |- (corridor -| adapter.south) -- (adapter.south);
\draw[arrow] (adapter.east) -- ++(0.45,0) |- ([yshift=0.18cm]unet.west);
\draw[arrow] (fpn.east) -- ++(0.45,0) |- ([yshift=-0.18cm]unet.west);
\draw[arrow] (unet.east) -- (head.west);
\draw[arrow] (head.east) -- (pmf.west);
\end{tikzpicture}%
}
\caption{Probability-backbone data flow. The frozen base reconstruction $\hat{X}=g(B)$ supplies both a scalar input to the 32-channel adapter and multi-scale features via the trainable 4-level FPN (45 injection sites), while the exact residual prefix and mask supply progressive context.}
\label{fig:f-backbone}
\end{figure}
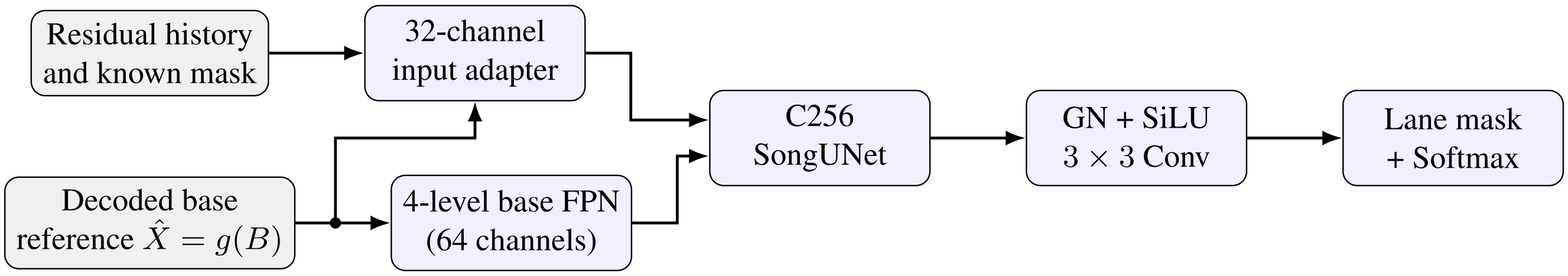

At stage $s$, a 32-channel input adapter (Table~\ref{tab:f-adapter}) concatenates the decoded token history, mask, global progress, lane identities, and packed base scalar, and feeds a four-scale C256 SongUNet (base width 256, channel multipliers $[1,2,2,2]$ across scales $256\times 768$ to $32\times 96$, 4 encoder and 5 decoder ResBlocks per scale, single-head self-attention in the first $32\times 96$ bottleneck block, 244.6M total parameters). In parallel, the normalized base tensor $\hat{\tau}/511$ passes through a trainable four-level 64-channel base feature pyramid whose features are projected via near-zero-initialized $1\times 1$ convolutions and added at 45 encoder/decoder sites (Figure~\ref{fig:f-backbone}). A shared head outputs 512 logits per token, masking invalid lane categories (including token $0$) prior to Softmax. Although the backbone originates from discrete diffusion models, progressive recovery is driven entirely by $(h^{(s)}, m^{(s)}, \rho_s)$ rather than a continuous noise schedule; accordingly, the time conditioning input is held fixed at $t_0=10^{-4}$ during both training and coding.

\begin{table}[tbp]
\centering
\caption{Composition of the 32-channel input adapter on a $256\times 768$ packed token plane.}
\label{tab:f-adapter}
\small
\renewcommand{\arraystretch}{1}
\begin{tabularx}{\linewidth}{lcY}
\toprule
Input component & Channels & Description\\
\midrule
Token embedding & 24 & Learned embedding over the 512-entry token vocabulary (including mask token $0$)\\
Token scalar & 1 & Decoded history token ID normalized by $511$\\
Known mask & 1 & Binary indicator: $1$ at decoded prefix positions ($G_{<s}$), $0$ elsewhere\\
Global progress & 2 & Spatially broadcast known and remaining fractions, $\rho_s$ and $1-\rho_s$\\
Lane identity & 3 & One-hot indicator of color lane (Y, Co, or Cg)\\
Base scalar & 1 & Packed reconstructed-base token value normalized by $511$\\
\midrule
Total & 32 & Concatenated along the channel dimension\\
\bottomrule
\end{tabularx}
\end{table}

\subsection{Lossless Entropy Coding and Exact Recovery Verification}\label{sec:impl-coding}
\enlargethispage{\baselineskip}

At each stage $s\in\{0,\ldots,47\}$, encoder and decoder run one \texttt{fp32} evaluation-mode forward pass (dropout disabled) to obtain the lane-masked PMFs over the packed plane, then sequentially encode or decode the $4{,}096$ symbols of $G_s$ in global row-major order using a 64-bit binary arithmetic coder with dynamic interval quantization and strict positive probability floors. The residual history $h^{(s)}$ and mask $m^{(s)}$ are updated only after all symbols in $G_s$ complete, preventing any within-group neural autoregression.

Exact reconstruction follows by induction across stages $s=0,\ldots,47$ whenever both endpoints (i) recover the exact integer reference $\hat{X}=g(B)$, (ii) execute the identical deterministic schedule $\pi$, and (iii) compute identical quantized coder intervals under matched arithmetic framing. After the final stage, Eq.~\eqref{eq:f-mod} and Eq.~\eqref{eq:f-inv} recover every RGB value identically. Every reported compression rate in the main paper and supplement---all $1{,}536$ encode--decode pairs across the six training and conditioning ablation configurations (Main Tables~4 and~5), all $1{,}024$ pairs in the four-way color-coordinate ablation (Main Table~1), all frozen-order and base-storage runs, and all $226$ original-resolution images ($9{,}168$ tiles)---was verified by full decode round-trips achieving exact pixel-wise RGB equality.

\end{document}